# ROBUST QUANTIFICATION OF PERCENT EMPHYSEMA ON CT VIA DOMAIN ATTENTION: THE MULTI-ETHNIC STUDY OF ATHEROSCLEROSIS (MESA) LUNG STUDY

*Xuzhe Zhang[1], Elsa D. Angelini[1,2], Eric A. Hoffman[3], Karol E. Watson[4], Benjamin M. Smith[5,6], R. Graham Barr[5], Andrew F. Laine[1]*

[1]Department of Biomedical Engineering, Columbia University, New York, NY, USA
[2]LTCI, Télécom Paris, Institut Polytechnique de Paris, Paris, France
[3]Department of Radiology, Medicine, and Biomedical Engineering, Univ. of Iowa, Iowa City, IA, USA
[4]Division of Cardiovascular Medicine, David Geffen School of Medicine, Los Angeles, CA, USA
[5]Department of Medicine, Columbia University Irving Medical Center, New York, NY, USA
[6]Department of Medicine, McGill University Health Center, Montreal, QC, Canada

## ABSTRACT

Robust quantification of pulmonary emphysema on computed tomography (CT) remains challenging for large-scale research studies that involve scans from different scanner types and for translation to clinical scans. Although the domain shifts in different CT scanners are subtle compared to shifts existing in other modalities (e.g., MRI) or cross-modality, emphysema is highly sensitive to it. Such subtle difference limits the application of general domain adaptation methods, such as image translation-based methods, as the contrast difference is too subtle to be distinguished. Existing studies have explored several directions to tackle this challenge, including density correction, noise filtering, regression, hidden Markov measure field (HMMF) model-based segmentation, and volume-adjusted lung density. Despite some promising results, previous studies either required a tedious workflow or eliminated opportunities for downstream emphysema subtyping, limiting efficient adaptation on a large-scale study. To alleviate this dilemma, we developed an end-to-end deep learning framework based on an existing HMMF segmentation framework. We first demonstrate that a regular UNet cannot replicate the existing HMMF results because of the lack of scanner priors. We then design a novel domain attention block, a simple yet efficient cross-modal block to fuse image visual features with quantitative scanner priors (a sequence), which significantly improves the results.

## 1. INTRODUCTION

Pulmonary emphysema is anatomically defined as the permanent enlargement of lung air spaces distal to the terminal bronchioles with the destruction of alveolar walls and parenchyma [1-3]. Emphysema and chronic obstructive pulmonary disease (COPD) jointly are the fourth leading cause of death in the US in 2019 [4].

Computed tomography (CT) is commonly used to quantify emphysema, which is usually represented by the fraction of voxels below a certain Hounsfield unit (HU) threshold within the full lung or as a percentile in the histogram distribution [5]. However, quantitative lung measures, including percent emphysema (*%emph*) and the 15th percentile (*perc15*), which is now being used as an endpoint in longitudinal cohort studies and clinical trials, is subject to scanner and protocol-specific biases [6, 7]. Specifically, *%emph* and *perc15* have been shown to be sensitive to factors that can influence the intensity distributions of lung CT scans (Fig. 1), including the reconstruction algorithm, slice thickness, scanner type, radiation dose, and inspiration level [3]. For large-scale studies that acquire longitudinal scans from different scanner types or protocols, spanning several years such as the Multi-Ethnic Study of Atherosclerosis (MESA) Lung study [8] and the Collaborative Cohort of Cohorts for COVID-19 Research (C4R) study [9], robust quantification of the emphysema remains challenging.

Previous studies have explored quantifying emphysema progression on heterogeneous full lung CT scans from different perspectives, including density correction based on air and aortic density [10], noise filtering [11, 12], a regression model [13], and hidden Markov measure field (HMMF) model-based segmentation [14-16]. Recently, calculating volume-adjusted lung density and using a linear mixed model to adjust for CT technical characteristics demonstrated promising results on evaluation of emphysema progression over 10 years in a multicenter study [17]. However, some previous explorations only tackled the challenge from a direction with intrinsic limitations. For example, [13] used a regression model to harmonize *%emph* measures across different reconstruction kernels but could not generate a valid emphysema mask for the downstream subtyping task [18, 19]. Similarly, this limitation also applies to the latest advancement [17]. The HMMF segmentation framework was reportedly superior to other methods by statistically modeling the intensity distributions of emphysematous and normal lung tissues and using them as priors to calculate the posterior probability of a voxel belonging to emphysema or not. However, the parametric model needs to be manually tuned when new scans come in. When applied to large-scale datasets, the manual intervention is inefficient and may be subject to operator bias.

Therefore, in this study, we aim to build a deep learning-based segmentation framework upon the HMMF method to achieve robust, fully automatic, and end-to-end quantification of emphysema that is scanner/protocol independent. We first demonstrate that the regular UNet model, despite its dominance and successful applications in the field of biomedical image segmentation, failed to replicate the HMMF masks on different CT scanner types, due mostly to the lack of scanner-specific priors in the model. Since pixels with similar textures and intensities will be determined as emphysema or non-emphysema based on different intensity distributions of distinct scanners, solely utilizing visual features as regular UNet will likely lead to poor generalization. Therefore, we aim to fuse the scan visual features with scanner priors that can be automatically derived to improve robustness. Inspired by the recent advancement in multi-modal learning, we designed a simple yet efficient domain attention module to fuse image features with another modality (i.e., the difference between cumulative density functions of a scanner and an individual scan, **Section 2.2**). We conducted comparisons between regular UNet and UNet with domain attention (UNet-DAttn) on in-distribution and out-of-distribution test sets and demonstrated that domain attention significantly improves both in-distribution accuracy and out-of-distribution generalization compared to the original HMMF masks.

## 2. MATERIALS AND METHODS

### 2.1. Data and CT Scanning

The Multi-Ethnic Study of Atherosclerosis (MESA) Study is a longitudinal, multicenter, and population-based prospective study to investigate subclinical cardiovascular disease [20]. MESA recruited 6,814 men and women ages 45-84 years old from six centers across the USA from 2000-02. The MESA Lung Study recruited 3,965 MESA participants in 2004-06. In 2010-12 as part of MESA Exam 5, 3,205 participants in the MESA Lung Study underwent full-lung CT scans with isotropic in-plane resolution varying from 0.4668 to 0.9180 mm and slice thickness of 0.625 or 0.75 mm following the MESA Lung CT protocol [21]. Lung segmentation was performed at a central reading center using VIDA software (VIDA Diagnostics, Inc. Coralville, IA, USA). The emphysematous tissue was segmented using HMMF as described in an earlier study [16], which produced binary emphysema masks ($emph_{HMMF}$) defining emphysema pixels as foreground class. The resultant percent emphysema was defined as $\%emph_{HMMF}$ and was calculated as the ratio of emphysema volume divided by full lung volume.

Four main CT scanner types were used in MESA Exam 5 from two manufacturers: LightSpeed VCT and Discovery STE (GE HealthCare, Chicago, IL USA) and Sensation 64 and Definition AS (Siemens Healthineers, Erlangen, Germany). We evaluated each type's performance for both in-distribution and out-of-distribution test scans to ensure its generalizability when applied to a new scanner unseen during training. For **training** and **validation**, we used scans acquired with GE Discovery STE, Siemens Sensation 64, and Siemens Definition AS scanners. We randomly sampled 100 full lung inspiratory scans per scanner type (n=300) as the training dataset (with average $\%emph_{HMMF}$ = 2.52, 3.90, 3.15, respectively; with max $\%emph_{HMMF}$ = 17.25, 29.89, 38.76, respectively), and 10 scans per scanner type (n=30) as the validation dataset (with average $\%emph_{HMMF}$ = 2.52, 1.81, 1.98, respectively; with max $\%emph_{HMMF}$ = 6.61, 4.84, 5.67, respectively). Per scan, we randomly sampled 50 axial slices within the full lung mask (n=15,000) for training, and 25 axial slices (n=750) for validation. For **in-distribution testing**, we used unseen 30 scans per scanner type (n=90) (with average $\%emph_{HMMF}$ = 3.35, 3.10, 2.76, respectively; with max $\%emph_{HMMF}$ = 21.95, 13.14, 10.68, respectively). For **out-of-distribution testing**, we used n=30 randomly sampled scans acquired with LightSpeed VCT (with average $\%emph_{HMMF}$ = 3.14 and max $\%emph_{HMMF}$ = 27.33). All axial slices containing lung were included in the test sets (n=47,793 and 15,357 for in-/out-of-distribution, respectively). While the scans-level $\%emph_{HMMF}$ varies from 0 to 38.76, slice-wise $\%emph_{HMMF}$ contain a larger range from 0 to 63.40.

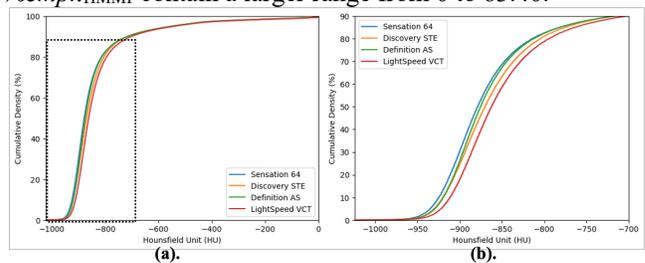

**Fig. 1.** (a) $CDF_{scanner}$ of Sensation 64, Discovery STE, Definition AS, and LightSpeed VCT. (b) Zoomed view of the dashed region in (a) (0-90% percentile and [-1024, -700] HU range).

### 2.2. Derivation of domain feature

As scan intensity distribution varied slightly with each scanner type and protocol, scans acquired from different scanner types with the same intensity distribution may contain different amounts of emphysematous tissue. In the previous HMMF-based methods [14-16], such distribution shifts were managed by adjusting posterior probabilities based on scanner-specific priors (i.e., parametric modeling of scanner intensity distribution).

To encode scanner-priors for emphysema appearance, we propose here to use scanner-specific cumulative density functions (CDF) (Fig. 1 and 2). This feature can be easily derived from the scans themselves. For each scanner type, we selected scans from non-diseased, never-smoking participants and computed the median of the scan-wise percentage of pixels below -950 HU (%-950). We selected 10 scans per scanner type around their median and concatenated all pixels within the lung region ($n_{pixels} \approx$ 200M). Scanner-specific $CDF_{scanner}$ were computed on 512 regular bin values (step size ~0.2). The four $CDF_{scanner}$ computed on MESA Lung Exam 5 are shown in Fig. 1. Some subtle differences in intensity distribution across different scanner types in the intensity range typically used for lungs ([-1024 -700] HU) can be observed in Fig. 1b.

Preliminary experiments suggested that using the $CDF_{scanner}$ as the domain feature was limited by its variance. Indeed, with only 3 training scanner types, low variance across $CDF_{scanner}$ seemed to lead to poor generalization when applied to a new scanner type. To tackle this limitation, we propose as an alternative feature to use the scan and scanner specific $CDF_{diff}$ computed as the difference between the CDF of individual scan ($CDF_{scan}$) and $CDF_{scanner}$ (Fig. 2).

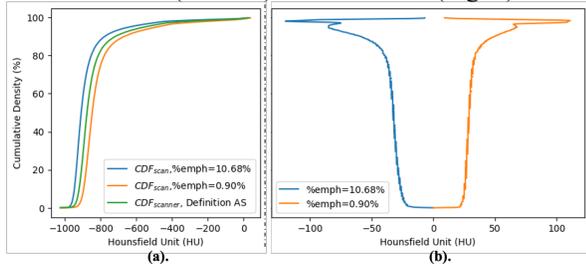

**Fig. 2.** Definition of CDF values: (a) $CDF_{scanner}$ of Definition AS and $CDF_{scan}$ of two scans acquired from Definition AS with %$emph_{HMMF}$ = 10.68% and 0.90%. (b) $CDF_{diff}$ = $CDF_{scan}$ - $CDF_{scanner}$ for the same two scans.

### 2.3. Model architecture and training

The overview of the proposed framework is depicted in Fig. 3. We adapted a regular UNet [22], which gradually downsamples spatial resolution from $512^2$ to $64^2$ and increases channel dimension from 1 to 512. We employed 2D convolution layers with stride of 2 and kernel size of 3 for downsamping (encoding) and used 2D transpose convolution with stride of 2 and kernel size of 4 for upsampling (decoding). A generic convolution block was used during encoding, consisting of group normalization (GN), a 2D convolution layer, and ReLU activation. When spatial dimension doesn't change, the convolution layer has a stride of 1 and a kernel size of 3. During decoding, we designed a double convolution block that contains 2x (GN+Conv2D+ReLU). The first convolution inside the double convolution block takes upsampled and skip-connection features as its input and decreases the channel number by one-half, while the second convolution layer does not change the channel dimension. All 2D convolution layers in double convolution blocks (Fig. 3a – green arrows) have a kernel size of 3 and a stride of 1.

To fuse image features and the $CDF_{diff}$, we designed a domain attention (DAttn) module, which was applied at each decoding layer. Inspired by the squeeze-and-excitation (SE) blocks [23], our DAttn employs fully connected layers, ReLU activation, and sigmoid activation for channel-wise attention (Fig. 3b). The $CDF_{diff}$ is first fed into a fully connected (FC) layer followed by ReLU activation and thereby encoded into a 256-D feature vector. This intermediate feature vector is then fed into another FC layer with an output channel equal to input image features. The output then goes through a Sigmoid activation and is converted to channel-wise attention weights. The difference between SE and DAttn is that SE takes the global average of image features as the attention weight vector, therefore being technically a channel-wise self-attention block, while the DAttn takes the image features and $CDF_{diff}$ as input to achieve domain attention.

$$L(y, \hat{y}) = -\sum_{i=1}^{H}\sum_{j=1}^{W} y_{i,j} \log \hat{y}_{i,j} - \frac{2\sum y\hat{y}}{\sum y + \sum \hat{y}} \quad (1)$$

UNet and DAttn are jointly trained to optimize a loss L combining cross entropy and Dice losses with equal weight (Eq. 1) where $y_{i,j}$ is the ground truth label as defined in $emph_{HMMF}$ mask and $\hat{y}_{i,j}$ is the predicted probability distribution. The learning rate was set as 0.0002 at the beginning of training and remained constant for the first 25 epochs. After 25 epochs, a cosine annealing scheduler with warm restart was used to decay the learning rate from $2e^{-4}$ to $1e^{-8}$, and then restart to $2e^{-4}$ periodically, where the first period takes 10 epochs, and the second period takes 20 epochs. The maximum number of epochs was set as 50 and we utilized a batch size of 8. After each epoch, validation was conducted and no improvement during validation for over 25 epochs would trigger early stopping. After training finished, the checkpoint with the best validation results was selected for testing. The AdamW optimizer was used with 0.05 weight decay and (0.9, 0.95) betas. A regular UNet (without DAttn), a UNet with DAttn using $CDF_{scanner}$, and a UNet with DAttn using $CDF_{diff}$ (proposed) were trained using the same configuration introduced above. When using $CDF_{scanner}$ for DAttn, to prevent training divergence and collapse caused by limited variance, a random noise N term was sampled from the uniform distribution U(-1,1) and added to $CDF_{scanner}$.

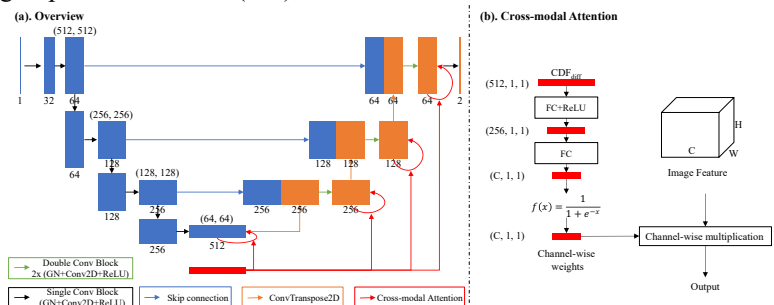

**Fig.3.** (a). Overview of the proposed UNet-DAttn. The number under each block indicates the channel dimension. (H, W) indicates the spatial resolution. (b). Illustration of the proposed domain attention block. FC = fully connected layer. GN = group normalization.

### 2.4. Evaluation

The framework was implemented using PyTorch. To evaluate the performance of the model, we use pixel-level Dice similarity coefficient (DSC) and scan-level signed error on emphysema measure defined as: *%emph*$_{HMMF}$ - *%emph*$_{pred}$. Results are reported at scan level for average values. To justify the limitation of using $CDF_{scanner}$ only, we conducted a comparative study on the choice of the domain CDF feature.

## 3. RESULTS

### 3.1. In-distribution and out-of-distribution test results

The models' performance on the in-distribution test set is reported in Table 1. The regular UNet achieved an average DSC of 66.71% and underestimated the *%emph* by 0.37% on average. The proposed DAttn improved the baseline performance in both mean error and average DSC while it overestimated the *%emph* by 0.21%.

**Table 1.** Performance of UNet and proposed UNet+DAttn on in-distribution test set, N = 90. Bold indicates the best results.

| Models | Metrics | | | |
|---|---|---|---|---|
| | *%emph*$_{HMMF}$ | *%emph*$_{pred}$ | Mean Error (%) | DSC |
| UNet | 3.07± 3.17 | 2.70± 2.02 | 0.37±1.42 | 66.71± 12.33 |
| UNet-DAttn (w/ $CDF_{diff}$) | | 3.28± 2.97 | **-0.21±1.27** | **70.23± 10.90** |

**Table 2.** Performance of UNet and proposed UNet+DAttn on out-of-distribution test set, N = 30. Bold indicates the best results.

| Models | Metrics | | | |
|---|---|---|---|---|
| | *%emph*$_{HMMF}$ | *%emph*$_{pred}$ | Mean Error (%) | DSC |
| UNet | 3.14± 5.13 | 2.30± 2.57 | 0.84±2.65 | 58.45 ±14.37 |
| UNet-DAttn (w/ $CDF_{scanner}$) | | 2.86± 3.36 | 0.27±1.90 | 60.66± 14.84 |
| UNet-DAttn (w/ $CDF_{diff}$) | | 2.98± 3.53 | **0.15±1.77** | **65.80 ±14.67** |

Results on out-of-distributions test scans are reported in Table 2. The regular UNet performed much worse than the in-distribution performance with decreased DSC and 127% larger mean error. For the UNet + DAttn, although the DSC also decreased compared to the in-distribution DSC, the mean error remained on the same level as the in-distribution test.

Performance of UNet and UNet+DAttn (with $CDF_{diff}$) at per scanner type level is reported in Fig. 4, where a more stable and tight distribution of the proposed method is observed.

We provide an example of visual segmentation results from different models on a LightSpeed VCT scan in Fig. 5. In this example UNet is prone to large false negative emphysema regions while UNet + DAttn shows scattered false positive pixels close to true-positive ones.

### 3.2. Selection of the domain feature

To justify selecting the $CDF_{diff}$ as one of the input features to DAttn, instead of $CDF_{scanner}$, we conducted an ablation experiment on the selection of domain feature. The results of models with different domain features were summarized in Table 2. Owing to the limited variance of $CDF_{scanner}$, the generalization of model was impaired compared to using $CDF_{diff}$ with richer variance. However, using $CDF_{scanner}$ as domain feature still improved over the baseline results of UNet.

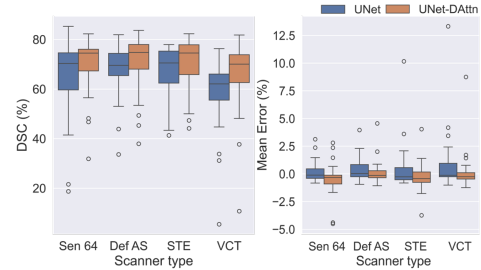

**Fig. 4**. Box plots of model's performance on scans acquired with different scanner types.

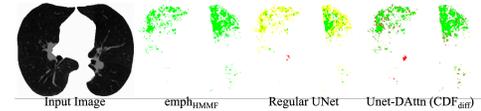

**Fig. 5**. Example of segmentation results from Regular UNet and UNet+DAttn. Green = true positive; yellow = false negative; red = false positive.

## 5. DISCUSSION

In this work, we propose a new deep learning framework using UNet segmentation with a novel domain attention mechanism to replicate the performance of a previously established HMMF-based segmentation of emphysema on multi-scanners lung CT scans, which eliminates potential operator bias and improves efficiency for large-scale application and deployment. The experiments on in-distribution and out-of-distribution test set, as well as one ablation study, demonstrate the effectiveness of the proposed domain attention mechanism. Moving forward, we plan to extend this framework to a 3D setting as emphysematous tissue have spatial coherence in 3D. Moreover, this work demonstrates promising preliminary results on a large-scale cohort collected at one timepoint. Application and analysis of the proposed framework on large-scale longitudinal datasets will be conducted as the next step.

## 6. COMPLIANCE WITH ETHICAL STANDARDS

This study was approved by the institutional review board and complied with HIPAA rules, and written informed consents were obtained from all participants.

## 7. ACKNOWLEDGMENTS

This research was supported by grants from the National Institutes of Health/National Heart, Lung, and Blood Institute (NIH/NHLBI) R01-HL121270, R01-HL077612, R01-HL093081, and R01-HL130506, and NIH/NHLBI contracts 75N92020D00001, HHSN268201500003I, N01-HC-95159, 75N92020D00005, N01-HC-95160, 75N92020D00002, N01-HC-95161, 75N92020D00003, N01-HC-95162, 75N92020D00006, N01-HC-95163, 75N92020D00004, N01-HC-95164, 75N92020D00007, N01-HC-95165, N01-HC-95166, N01-HC-95167, N01-HC-95168 and N01-HC-95169, and by grants UL1-TR-000040, UL1-TR-001079, and UL1-TR-001420 from the National Center for Advancing Translational Sciences. The authors thank the other investigators, the staff, and the participants of the MESA study for their valuable contributions. A full list of participating MESA investigators and institutions can be found at http://www.mesa-nhlbi.org.

Author E.A.H is a shareholder of VIDA Diagnostics, Inc.